\documentclass{article}
\usepackage[utf8]{inputenc}
\usepackage[auth-sc,affil-it]{authblk}
\usepackage{arydshln} 

\title{Do Multilingual Language Models Capture Differing Moral Norms?}

\author[1]{Katharina Hämmerl}
\author[2,3]{Björn Deiseroth}
\author[3]{Patrick Schramowski}
\author[5,1]{Jindřich Libovický}
\author[1]{Alexander Fraser}
\author[3,4]{Kristian Kersting}
\affil[1]{Center for Information and Language Processing, LMU Munich, Germany}
\affil[2]{Aleph Alpha GmbH, Heidelberg, Germany}
\affil[3]{Artificial Intelligence and Machine Learning Lab, TU Darmstadt, Germany}
\affil[3]{Hessian Center for Artificial Intelligence (hessian.AI), Darmstadt, Germany}

\affil[5]{Institute of Formal and Applied Linguistics, Charles University, Czech Republic}
\date{}

\usepackage[numbers]{natbib}
\usepackage{graphicx}
\usepackage{xcolor}

\begin{document}

\maketitle

\begin{abstract}
Massively multilingual sentence representations are trained on large corpora of uncurated data, with a very imbalanced proportion of languages included in the training. This may cause the models to grasp cultural values including moral judgments from the high-resource languages and impose them on the low-resource languages. The lack of data in certain languages can also lead to developing random and thus potentially harmful beliefs. Both these issues can negatively influence zero-shot cross-lingual model transfer and potentially lead to harmful outcomes. Therefore, we aim to
(1) detect and quantify these issues by comparing different models in different languages, (2) develop methods for improving undesirable properties of the models.
Our initial experiments using the multilingual model XLM-R show that indeed multilingual LMs capture moral norms, even with potentially higher human-agreement than monolingual ones. 
However, it is not yet clear to what extent these moral norms differ between languages.
\end{abstract}



Recent work demonstrated large pre-trained language models (PLM) obtain symbolic, relational \cite{petroni2019language} but also commonsense knowledge \cite{davison19commonsense}.
Further, West \textit{et al.} \cite{west21symbolic} showed that one is able to extract the commonsense knowledge from the large, general language model GPT-3 \cite{brown20language} via symbolic knowledge distillation. 
This encoded ``knowledge'' includes information of our society reflecting ethical and social norms \cite{jentzsch19semantics, schramowski22language}. Hereby the knowledge of a PLM is acquired during the self-supervised pre-training phase, which in case of most current state-of-the-art models uses scraped data from the web.

Therefore, approaches investigating the agreement of the model's norms and human values \cite{schramowski22language} and benchmark datasets \cite{hendrycks21aligning} aiming to align human and machine values with human labelled data arose. The work of Jiang \textit{et al.}~\cite{jiang21delphi} showed promise in terms of obtaining such alignment, i.e., teaching these kinds of models commonsense moral reasoning.


However, social norms are constantly evolving and differ between cultures. Whereas ``general'' alignment is an open question, teaching AI systems moral norms includes the representation of (moral) values from different societies, e.g., cultures. Can one model differentiate between cultural differences or can we observe differences in moral norms in models trained on different cultural data in the first place? The results of \cite{schramowski2020themoral} show promise for at least the latter question.

\paragraph{Can multilingual language models capture moral norms?}
Multilingual language models are trained on large corpora of uncurated data, with a very imbalanced proportion of languages included in the training. While basic semantic properties are often accessible across languages, there is an inherent problem in achieving perfect language neutrality \cite{libovicky20:findingsemnlp}, meaning that two sentences in differing languages with the same semantics receive very similar embeddings. There are techniques for improving the alignment, for instance, we have worked on an approach combining the strengths of static and contextualised embeddings \cite{haemmerl22:findingsacl}. But a very interesting and unexplored area of research is to consider whether multilingual language models capture differing moral norms, e.g., that the moral norms corresponding to the Chinese space in a multilingual language model may systematically differ from those in the Portuguese space.

\paragraph{Research Questions.}
With this study, we build upon these recent findings and aim to investigate the following:
\begin{itemize}
    \item The extraction of commonsense knowledge on moral norms from pre-trained language models.
    \item Does a multilingual model encode differing information compared to monolingual models?
    \item Whether a multilingual language model can capture differing moral norms from several language sources.
\end{itemize}

\begin{table}
\centering
\begin{tabular}{lccccc}
\hline
\textbf{Model} & \textbf{en} & \textbf{ar} & \textbf{cs} & \textbf{de} & \textbf{zh} \\ \hline
mBERT (mean-pooled) & 0.65 & -0.11 & -0.11 & -0.21 & 0.61 \\
XLM-R (mean-pooled) & -0.15 & -0.03 & 0.01 & -0.17 & -0.01 \\
monolingual (mean-pooled) & -0.15 & 0.43 & 0.01 & 0.15 & 0.69 \\
\hdashline
monolingual BERT (S-BERT) \cite{schramowski22language} & 0.79 &---&---&---&--- \\
XLM-R (best S-BERT) & 0.85 & 0.82 & 0.84 & 0.82 & 0.80 \\
\hline
\end{tabular}
\caption{Initial experiments with different multilingual and monolingual transformer models in the \textsc{MoralDirection} framework.
The multilingual XLM-R model achieves a higher correlation with human moral norms than the monolingual BERT.
}
\label{tab:initial-multiling-md}
\end{table}

From initial experiments, a version of XLM-R tuned with the S-BERT framework \cite{reimers-gurevych-2020-making}\footnote{The best correlations were achieved by \texttt{sentence-transformers/xlm-r-100langs-bert\-base-nli-mean-tokens}} shows good correlation with the global user study conducted by \cite{schramowski22language} when used with their \textsc{MoralDirection} (MD) framework (see Table~\ref{tab:initial-multiling-md}).
Simply mean-pooling representations from XLM-R \cite{conneau-etal-2020-unsupervised}, mBERT \cite{devlin-etal-2019-bert}, or monolingual Transformer models \cite{devlin-etal-2019-bert, antoun2020arabert, straka-etal-2021-robeczech, chan-etal-2020-germans} generally does not achieve a correlation, highlighting the need for semantic sentence representations for this goal.
There are some exceptions to this rule which may be due to details in how the different models are trained and how much training data is available for each language in the multilingual models.
Since this specific version of XLM-R was tuned with some parallel data, and these numbers were obtained from a single user study with a relatively small set of moral statements, it is difficult to say how much this reflects shared moral norms between the respective cultures, and
to what extent
it reflects the internal alignment of the model.

In Table~\ref{tab:correlation_xlm-r} we can observe
very high alignment between languages within the XLM-R model, which may not be surprising. Recall that the model is trained so that two sentences in different languages with the same semantics receive very similar embeddings. Also note that the tested statements provided by \cite{schramowski22language} do not aim to grasp cultural differences. An interesting question raised by these initial results is whether
language alignment
is in fact
desirable 
when considering
moral norms, which can differ in differing cultures.
\begin{table}[]
    \centering
    \begin{tabular}{l|ccccc}

\textbf{language} &\textbf{en} & \textbf{ar}&\textbf{cs}&\textbf{de}&\textbf{zh}\\\hline
\textbf{en}&1.0&---&---&---&---\\
\textbf{ar}&0.92&1.0&---&---&---\\
\textbf{cs}&0.93&0.96&1.0&---&---\\
\textbf{de}&0.94&0.96&0.98&1.0&---\\
\textbf{zh}&0.92&0.96&0.95&0.96&1.0\\
    \end{tabular}
    \caption{Correlation of XLM-R languages, \textit{cf.} Table~\ref{tab:initial-multiling-md}.}
    \label{tab:correlation_xlm-r}
\end{table}

Summarised, these observations already confirm the results of \cite{schramowski22language} in a larger multilingual setting and indicate that multilingual LMs indeed capture moral norms. To what extent they differ, however, is still unclear.
Therefore, we further aim to clarify this by experimenting on monolingual as well as multilingual transformer models.
\bibliography{references}
\end{document}